\documentclass[times,twocolumn,final,authoryear]{elsarticle}

\usepackage{ycviu}
\usepackage{framed,multirow}

\usepackage{amssymb}
\usepackage{latexsym}

\usepackage{url}
\usepackage{xcolor}
\definecolor{newcolor}{rgb}{.8,.349,.1}

\usepackage{hyperref}
\usepackage{amsmath,amssymb,amsfonts}
\usepackage{algorithmic}
\usepackage[ruled,longend]{algorithm2e}
\usepackage{graphicx}
\usepackage{textcomp}

\usepackage{amsmath}
\usepackage{multirow}
\usepackage{multicol}

\usepackage{lipsum}

\newcommand\blfootnote[1]{%
  \begingroup
  \renewcommand\thefootnote{}\footnote{#1}%
  \addtocounter{footnote}{-1}%
  \endgroup
}

\usepackage{amsmath,amsfonts,bm}

\def\eqref#1{equation~\ref{#1}}
\def\1{\bm{1}}

\def\ra{{\textnormal{a}}}
\def\rb{{\textnormal{b}}}

\def\ro{{\textnormal{o}}}

\def\ru{{\textnormal{u}}}

\def\rz{{\textnormal{z}}}

\def\va{{\bm{a}}}

\def\vc{{\bm{c}}}

\DeclareMathAlphabet{\mathsfit}{\encodingdefault}{\sfdefault}{m}{sl}
\SetMathAlphabet{\mathsfit}{bold}{\encodingdefault}{\sfdefault}{bx}{n}

\DeclareMathOperator*{\argmax}{arg\,max}

\newcommand{\pickObjectAttack}{\textsc{Pick-Object-Attack}\xspace}
\newcommand{\UnTargetedAll}{\textsc{DAG}\xspace}

\newcommand{\TargetedConf}{\textsc{Tar-Confident}\xspace}
\newcommand{\TargetedFreq}{\textsc{Tar-Frequent}\xspace}
\newcommand{\UnTargetedConf}{\textsc{Non-Tar-Confident}\xspace}
\newcommand{\UnTargetedFreq}{\textsc{Non-Tar-Frequent}\xspace}
\DeclareMathOperator{\SSIM}{SSIM}

\newcommand{\norm}[1]{\left\lVert#1\right\rVert}

\newcommand{\opick}{\ensuremath{\ro_{pick}}\xspace}
\newcommand{\classK}{\ensuremath{k}\xspace}
\newcommand{\green}[1]{\textcolor{black}{#1}}
\newcommand{\newgreen}[1]{\textcolor{black}{#1}}

\journal{Computer Vision and Image Understanding}

\begin{document}

\ifpreprint
  \setcounter{page}{1}
\else
  \setcounter{page}{1}
\fi

\begin{frontmatter}

\title{\pickObjectAttack: Type-Specific Adversarial Attack for Object Detection}

% \author[1]{Given-name \snm{Surname}\corref{cor1}} 
% \cortext[cor1]{Corresponding author: 
%   Tel.: +0-000-000-0000;  
%   fax: +0-000-000-0000;}
% \ead{author@author.com}
% \author[2]{Given-name \snm{Surname}}
% \author[2]{Given-name \snm{Surname}}

% \address[1]{Affiliation 1, Address, City and Postal Code, Country}
% \address[2]{Affiliation 2, Address, City and Postal Code, Country}

\author[mymainaddress]{ Omid \snm{Mohamad Nezami}\corref{mycorrespondingauthor} \fnref{contribution}}
\cortext[cor1]{Corresponding author}
\ead{omid.mohamad-nezami@hdr.mq.edu.au}

\author[mysecondaryaddress]{Akshay \snm{Chaturvedi} \fnref{contribution}}
\fntext[contribution]{The authors contributed equally to this work.}

\author[mymainaddress]{Mark \snm{Dras}}
\author[mysecondaryaddress]{Utpal \snm{Garain}}

\address[mymainaddress]{Macquarie University, Sydney, NSW, Australia}
\address[mysecondaryaddress]{Indian Statistical Institute, Kolkata, India}

\received{1 May 2013}
\finalform{10 May 2013}
\accepted{13 May 2013}
\availableonline{15 May 2013}
\communicated{S. Sarkar}

\begin{abstract}
Many recent studies have shown that deep neural models are vulnerable to adversarial samples: images with imperceptible perturbations, for example, can fool image classifiers. In this paper, we present 
% an adversarial attack against
the first type-specific approach to generating adversarial examples for object detection, which entails detecting bounding boxes around multiple objects present in the image and classifying them at the same time, making it a harder task than against image classification.  We specifically aim to attack the widely used Faster R-CNN by changing the predicted label for a particular object in an image: where prior work has targeted one specific object (a stop sign), we generalise to arbitrary objects, with the key challenge being the need to change the labels of \emph{all} bounding boxes for all instances of that object type. To do so, we propose a novel method, named \pickObjectAttack. \pickObjectAttack successfully adds perturbations only to bounding boxes for the targeted object, preserving the labels of other detected objects in the image. In terms of perceptibility, the perturbations induced by the method are very small. Furthermore, for the first time, we examine the effect of adversarial attacks on object detection in terms of a downstream task, image captioning; we show that where a method that can modify all object types leads to very obvious changes in captions, the changes from our constrained attack are much less apparent.
\end{abstract}


\end{frontmatter}

%\linenumbers
\section{Introduction}

Deep learning systems have achieved remarkable success for several computer vision tasks. However, adversarial attacks have brought into question the robustness of such systems.  \cite{goodfellow2014explaining} and \cite{szegedy2013intriguing} presented early attacks against image classifiers, using gradient-based techniques to construct inputs with the ability to fool deep learning systems. Since then adversarial attacks have been extensively studied for image classification, including being shown to be transferable across different image classifiers \citep{liu2017delving}. 
This presents risks for many applications of image processing, such as in forensics or biometrics;\footnote{There is in fact a contemporaneous special issue of this journal on this topic \citep{chellappa-etal:2021:CVIU}.} \blfootnote{© 2021. This manuscript version is made available under the CC-BY-NC-ND 4.0 license https://creativecommons.org/licenses/by-nc-nd/4.0/} for example,  \cite{caldelli-etal:2019:SPIC} present a collection of work on image and video forensics, such as locating splicing forgeries in images (tampered areas of synthesized images) \citep{liu-pun:2018:SPIC} or detecting image inpainting \citep{zhu-etal:2018:SPIC}. 
Adversarial attacks against images are usually categorised into two types (i) Targeted and (ii) Non-targeted. In a targeted attack, the goal is to modify the input so as to make the deep learning system predict a specific class, whereas in a non-targeted attack, the input is modified so as to cause the prediction of any incorrect class.

\begin{figure*}[!t]
\begin{center}
% \fbox{\rule{0pt}{2in} \rule{.9\linewidth}{0pt}}
\includegraphics[width=0.7\linewidth]{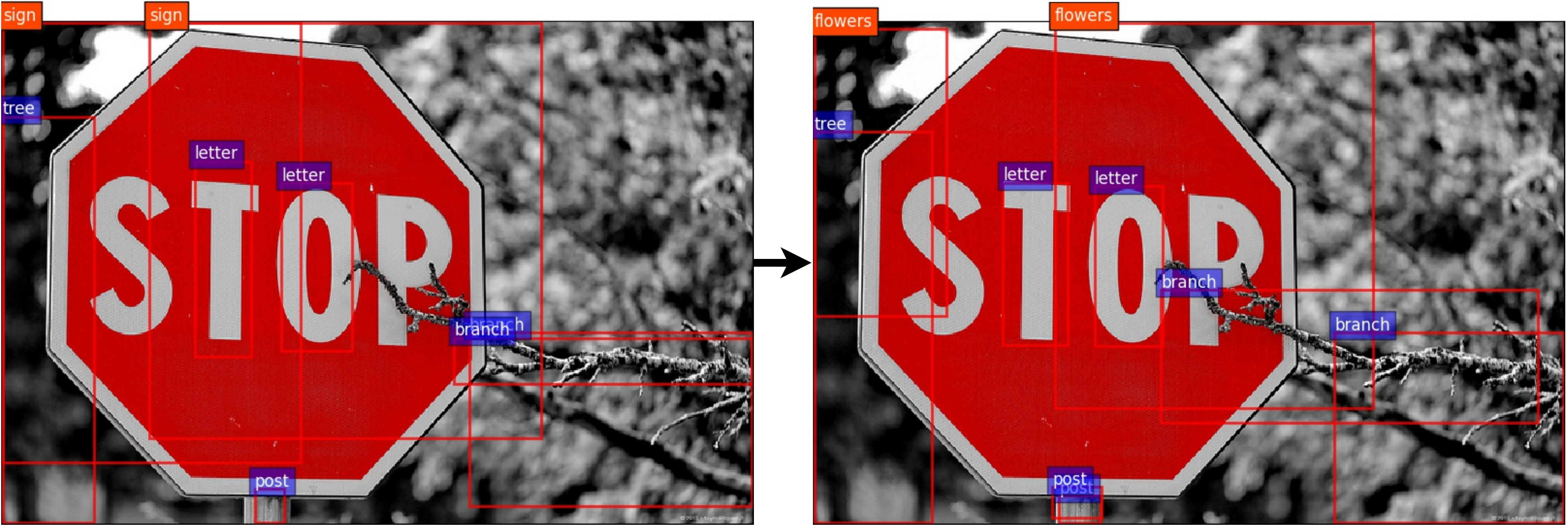}
\end{center}
   \vspace{-5mm}
   \caption{Example of our adversarial attack. \pickObjectAttack adds imperceptible perturbations to the first image (on the left) 
   resulting in the second image (on the right). It succeeds in changing the predicted class of the targeted object from ``sign'' to ``flowers'' (shown in orange) while other predicted classes (shown in blue) are unchanged.}
\label{fig:example_intro}
\vspace{-5mm}
\end{figure*}

A more challenging task is to construct adversarial examples that will fool an object detection system, with each image containing multiple objects and multiple proposals for each object; \cite{xie2017adversarial} provide an analysis of this complexity. \cite{chen2018shapeshifter} motivate this task with the example of object detection by an autonomous vehicle to recognise a stop sign and the risks involved in an adversarial attack in that context.

These two works tackle the issue of adversarial attacks against object detection and are the most relevant to our work. \cite{xie2017adversarial} propose a non-targeted attack where the predictions of all objects are changed simultaneously. \cite{chen2018shapeshifter} propose an attack against the object detector to misclassify only stop sign images; the attack method deliberately adds perceptible noise to the images.

In this paper, the proposed \pickObjectAttack aims to change the label of a particular object while keeping the labels of other detected objects unchanged. In this sense, it is a generalisation of \cite{chen2018shapeshifter}, where there may be a particular object that the attacker wants to be misclassified. More generally, it is often a goal of adversarial attacks to be imperceptible to observers; attacking just a single object, with the small number of bounding boxes involved, minimises the changes to the image relative to modifying all the objects as in \cite{xie2017adversarial}.  Moreover, changes to the image --- even if imperceptible to humans --- could be perceptible via downstream tasks.  For instance, object detection plays a crucial role in the state-of-the-art visual question answering (VQA) and image captioning systems \citep{anderson2018bottom}. Changing the entire image may lead to dramatically different answers or captions and hence alert the user indirectly.
Figure~\ref{fig:example_intro} shows an example of our proposed \pickObjectAttack. From the figure, we can see that only the label of object type ``sign'' has changed to ``flowers'' whereas other objects are detected correctly. This is because the perturbation is only added to the bounding boxes with the predicted label ``sign''.
%Figure  \ref{fig:example_intro} shows an example of our proposed attack, where only the label of a particular object is changed from ``sign'' to ``flowers'' whereas other objects are detected correctly. This is because the perturbation is only added to the bounding boxes with the predicted label ``sign''.
Analysing perceptibility of this attack both by standard metrics and via a downstream task contributes to the understanding of risks for forensics.

In this paper, we propose both targeted and non-targeted versions of \pickObjectAttack against Faster R-CNN \citep{ren2015faster}, a widely used and high-performing object detector, which \cite{chen2018shapeshifter} and \cite{xie2017adversarial} also based their work on. Where they
studied a version of Faster R-CNN trained on the MSCOCO dataset \citep{lin2014microsoft}, we use a different Faster R-CNN trained on the Visual Genome dataset \citep{krishna2017visual}, covering a larger set of classes than the MSCOCO dataset. Bottom-up features obtained from this version of Faster R-CNN have been employed in state-of-the-art VQA and image captioning systems \citep{anderson2018bottom}. These systems use the bottom-up and top-down attention to attend to the bounding boxes in order to generate a caption (or answer).

The main contributions of this paper are as follows \footnote{Our source code is publicly available at the following link:

\url{https://github.com/omidmnezami/pick-object-attack}}:

\begin{enumerate}[I]
\item 
This is the first study to successfully apply both targeted and non-targeted attacks against Faster R-CNN on different types of images. \cite{xie2017adversarial} only study the non-targeted attack against Faster R-CNN and \cite{chen2018shapeshifter} only attack stop sign images. The proposed attack achieves high success rates for both targeted ($>75\%$) and non-targeted ($>95\%$) attacks. We show that the proposed attack adds imperceptible perturbation to the image. \newgreen{Furthermore, the perturbation is only added to the pixels within a specified object type.}

\item
% This is the first work which studies an adversarial attack against Faster R-CNN in a constrained setting where only pixels within a specified object type are changed. Our proposed attack changes the label of a particular object in an image while preserving the labels of other detected objects. We propose an attack which works for arbitrary images and can be straightforwardly generalised to change the labels of multiple detected objects.
\newgreen{This is the first work which studies an adversarial attack against Faster R-CNN in a constrained setting where the label of a particular object in an image is changed while preserving the labels of other detected objects which are outside of the region of interest. We propose an attack which works for arbitrary images and can be straightforwardly generalised to change the labels of multiple detected objects.}

\item
This is the first work to use a downstream task to contribute to an understanding of perceptibility of adversarial attacks.  Specifically, we study the effect of attacking Faster R-CNN on the state-of-the-art image captioning system \citep{anderson2018bottom} which uses bottom-up features.  We show that it leads to many fewer changes in captions than a method based on \cite{xie2017adversarial} which modifies all the objects.
%Comprehensive analyses show that our proposed attack can easily fool the captioning model.
\end{enumerate}

% The success rate of the proposed attack is as high as the work of Chen \cite{chen2018shapeshifter}, even though the proposed attack adds imperceptible noise and targets different types of images.
\vspace{-6mm}
\section{Related work}
\vspace{-3mm}
% In this section, we give a brief overview on adversarial attack for different vision tasks and discuss related work on adversarial attack against Faster R-CNN in detail.

% \subsection{Adversarial Attack}
The generation of adversarial samples was first investigated in the context of deep learning by \cite{szegedy2013intriguing}, who used a gradient-based optimization to arbitrarily manipulate the input sample of a deep neural network for image classification. This manipulation usually aims to find similar samples with  differences that are imperceptible to human observers, in order to change the predicted class. Later works \citep{nguyen2015deep,carlini2017towards,eykholt2018designing} have led to better methods for generating adversarial samples, via different attack mechanisms, to mislead different classification models. In addition to classification, adversarial samples have also been crafted for other tasks such as image captioning \citep{chen2018attacking,ji2020attacking} and \newgreen{visual question answering \citep{chaturvedi2020attacking}}. They studied earlier image captioning models \citep{vinyals2015show,xu2015show} which use features from image classifiers. 

However, in many scenarios including the physical world, we usually face multiple objects in an image. Under such a condition, an attack would be required to fool an object detector, which detects the bounding boxes of objects in addition to classifying them. \cite{eykholt2018robust} discussed how misleading an object detector, such as YOLO~\citep{Redmon_2017_CVPR} and Faster R-CNN~\citep{ren2015faster}, is more difficult than misleading an image classifier.

% In this paper, we attack Faster R-CNN, which is a widely used and high-performing system for object detection. 
\green{In this paper, we propose \newgreen{a white-box attack against }Faster R-CNN, which is a highly-cited and high-performing system for object detection. It has been recently used for different important purposes such as object tracking and segmentation \citep{wang2019fast,chen2019tensormask}, image captioning and visual question answering \citep{anderson2018bottom,Gao_2019_CVPR} and so forth.}
\newgreen{Faster R-CNN consists of two stages, a region proposal network (RPN) for detecting the bounding boxes of objects, and a classifier for classifying the boxes \citep{ren2015faster}. Although the RPN can generate a dynamic number of bounding boxes from the image, an upper bound is usually set on the number of bounding boxes ranked by their confidence levels. The confidence level of each bounding box is calculated using the objectness score and non-maximum suppression (NMS). The RPN predicts an objectness score indicating the probability of an object being present inside the box and the NMS threshold reduces the number of detected boxes. The output of Faster R-CNN will be the classification for the detected boxes.}

% The purpose of our adversarial attack on Faster R-CNN is to change the prediction $c_i$ for specific detected object in multiple bounding boxes.

%% MD 190910
%% - from Chen et al, add citations [13] and [7] and discuss briefly.

\newgreen{\cite{xu2020adversarial} analysed several popular attacks against object detection models including Faster R-CNN and YOLO. } \cite{chen2018shapeshifter} proposed both \textit{targeted} and \textit{non-targeted} attacks on Faster R-CNN but only for stop sign images. They selected stop signs due to security-related issues in the real world, e.g. self-driving cars. They added \textit{perceptible} perturbations to make their adversarial samples robust after printing. 
Very recently, \cite{huang2019attacking} targeted stop signs, but by adding \textit{perceptible} perturbations around the border of the signs.
In contrast, we target a random set of different objects for both \textit{targeted} and \textit{non-targeted} attacks. We add \textit{imperceptible} perturbations to fool Faster R-CNN.
% ; these perturbations do not change all pixel values of the original samples
\cite{xie2017adversarial} proposed a \textit{non-targeted} attack \newgreen{(DAG)} on Faster R-CNN. They added \textit{imperceptible} perturbations to \textit{all pixels} in the input image to change the classes for all detected objects. Here, for the adversarial image, the RPN usually generates a different set of bounding boxes, with different scales. The bounding boxes change because adding the perturbations can change their confidence levels. They also changed the upper bound of detected boxes from $300$ to $3000$, guaranteeing the transfer of classification error among nearby boxes. \newgreen{The DAG approach has been applied on the PASCAL VOC dataset containing $21$ classes. \cite{wang2020adversarial} also proposed a \textit{non-targeted} attack on Faster R-CNN and compared against the DAG approach. However, they only used 256 proposals rather than original 3000 ones in \cite{xie2017adversarial} and did their experiment on Pascal VOC dataset. The notion of success cases in their work is different from \cite{xie2017adversarial}. They considered their attacks successful where either the labels or the location of original bounding boxes are changed or new bounding boxes are introduced. However, \cite{xie2017adversarial} called an attack successful if  none of the original classes is detected. Thus, we compare against the DAG approach of \cite{xie2017adversarial}.} In our \pickObjectAttack, we do not increase the upper bound of number of boxes and only add \textit{imperceptible} perturbations to the boxes corresponding to a targeted object to change its predicted class. We do not change the pixel values of other boxes. \newgreen{Unlike \cite{xie2017adversarial} and \cite{wang2020adversarial},} we study both \textit{targeted} and \textit{non-targeted} attacks. \newgreen{Concurrent with our work, \cite{chow2020understanding} proposed a targeted attack in three different settings including object-vanishing, object-fabrication and object-mislabeling. They performed their experiments on Pascal VOC dataset and MSCOCO dataset containing $81$ classes. In contrast, we apply our attack on the Visual Genome dataset containing $1600$ classes, and as with previous work, our task definition of attacking all instances of a particular object type differs.} For the first time, we also examine the impact of our proposed attacks on the state-of-the-art image captioning system \citep{anderson2018bottom}.

\vspace{-2mm}
\section{Method}
\vspace{-3mm}
\subsection{Faster R-CNN Model}
We evaluate our attack method against Faster R-CNN with ResNet-101, pre-trained on the ImageNet dataset \citep{deng2009imagenet}, then trained on the object and attribute instances of the Visual Genome dataset. The model leads to the state-of-the-art on different tasks like image captioning and visual question answering \citep{anderson2018bottom} in addition to generating a high object detection performance. Previous works \citep{xie2017adversarial,chen2018shapeshifter} studied attacking Faster R-CNN trained on the MSCOCO dataset \citep{lin2014microsoft} having only $80$ object classes. In comparison, the Visual Genome dataset has $1600$ object classes. It includes $3.8$M object instances versus $1.5$M for MSCOCO. It also contains $2.8$M attributes and $2.3$M relationships.
% \footnote{https://github.com/peteanderson80/bottom-up-attention}.
% It has two stages as in the original Faster R-CNN \cite{ren2015faster}. In the first stage, it uses a region proposal network (RPN) to detect object bounding boxes. In the second stage, it extracts the features of each bounding box to generate its classification score. In the first stage, the top bounding boxes are usually chosen; these are then fed into the second stage, using greedy non-maximum suppression (NMS) with a confidence threshold. The confidence level of each bounding box is calculated using the intersection-over-union (IoU).
% and select the top bounding box to attack.

% \subsection{\pickObjectAttack}
\subsection{PICK-OBJECT-ATTACK}
Let $I_{org}$ denote the original image. Let $N$ be the number of bounding boxes and $K$ be the number of classes. An object detector can be mathematically expressed as a function $f: I_{org} \longrightarrow (g, h)$ where $g \in \mathbb{R}^{N\times K}$ denotes the probability distribution for $N$ bounding boxes, and $h \in \mathbb{R}^{N\times 4}$ denotes the predicted coordinates of the bounding boxes. Let \opick denote the selected object to attack and $\va\subseteq \{1,2,..,N\}$ denote the indexes of the boxes with predicted class \opick for the image $I_{org}$. 
% I_{org} with shape s. goal is to get I_{adv} with same shape s. Faster R-CNN reshapes the images to make shortest side 600. I_{org}' reshaped original image. 
Faster R-CNN rescales the input image so that the shortest size is $600$ pixels. Given the original image $I_{org}$ of shape $s$, our proposed attack generates an adversarial image $I_{adv}$ of the same shape. 
% This is in contrast with previous attacks which generate adversarial image with shape corresponding to the input of Faster R-CNN (check this).
For an image $I$, we denote the rescaled image (with the shortest side being $600$) by $I'$. 

\paragraph{Mask Detection}
% Given I_{org} and an object o, prepare mask of shape s.
As mentioned before, our proposed attack aims to change the label of a targeted object \opick. To do so, for each attack, we prepare a binary mask denoted by $M$ which has a same shape as $I_{org}$. $M$ is $1$ for bounding boxes with predicted label \opick and is $0$ otherwise. We compute $M$ for the original image and keep $M$ fixed during the attack.

For a loss function $L$ and an image $I'$ (obtained by rescaling image $I$), we obtain $\nabla_{I'}' L$ during the \emph{backward pass} given by 
\begin{equation}
    \nabla_{I'}' L = r \nabla_{I'} L
\label{eq:grad}
\end{equation}

\noindent
where $r$ is the learning rate and $\nabla_{I'} L$ is the gradient of loss $L$ for image $I'$. We resize the gradient $\nabla_{I'}' L$ and apply the mask $M$ according to the following equation

\begin{equation}
    \nabla_{I} L = M \odot \text{rescale}(\nabla_{I'}' L,s)
\label{eq:grad_org}
\end{equation}

\noindent
where $\odot$ denotes bitwise multiplication. For the proposed attack, we use the final obtained gradient $\nabla_{I} L$ for updating image $I$. 
% Backward pass:
%0. Forward pass with I'
%1. calculate gradient grad with respect to I'
%2. grad = grad*lr
%3. grad = resize(grad,s)
%4. grad = grad*m
%5. Update I.
%6. resize I to I'
%7. check stopping criteria for I' 
We explain the loss functions for both the non-targeted and targeted attacks below. 

% and $c_{a}$ denotes the predicted classes for the attack boxes.

% Our proposed attack uses the following loss function, $L$ given by

% \begin{equation}
%     L = - log(g_{a,c_{a}})
% \label{eq:loss}
% \end{equation}

% where $g_{i,j}$ denotes the predicted probability of the $j^{th}$ class for the $i^{th}$ box. The proposed attack modifies the image $I$, via gradient-ascent, using the gradient, i.e. $\nabla_{I}' L$ given by

% \begin{equation}
%     \nabla_{I}' L = r \nabla_{I} L
% \label{eq:grad}
% \end{equation}
%  where $\nabla_{I}L$  is the gradient of loss $L$ with respect to the image I and $r$ is the learning rate. We study both targeted and non-targeted attacks explained in the following sections.

% \begin{algorithm}[!h]
% \caption{\UnTargetedConf / \UnTargetedFreq}\label{alg:non-targeted}
% \DontPrintSemicolon
%   $\textbf{Input:}I_{org}, r, max_{iter}, \opick$\;
%   $\textbf{Output:}I_{adv}$\;
%   $\text{Get mask }M\text{ from }I_{org}\text{ and } \opick$\;
%   $success \gets \textbf{False}$\;
%   $I \gets I_{org}$\;
%   \For{$j\leftarrow 1$ \KwTo $max_{iter}$}
%   {
%   $\text{Compute } \va \text{ for image } I$\; 
%   \uIf { $ \va = \varnothing$} %$\ro_{pick} \neq \argmax\limits_{c} g_{\ra,c} \text{ for any } \ra \in \va$}
%   {
%   $success \gets \textbf{True}$\;
%   $\textbf{break}$\;
%   }
%   $\text{Compute loss } L \text{ using }\eqref{eq:loss}$\;
%   $\text{Compute }  \nabla_{I} L \text{ using }\eqref{eq:grad_org}$ \;
%   $I \gets I + \nabla_{I} L$\;
%   $\text{Truncate image } I \text{ in the range } [0,255]$\; 
%   }
%   $I_{adv} \gets I$\;
%   \KwRet{$I_{adv}$}\;
% \end{algorithm}

 \paragraph{Non-Targeted Attack}
 Our goal in the non-targeted attack is to generate an image $I_{adv}$ so that none of the detected boxes have the predicted class $\ro_{pick}$. To achieve this, we use the following loss function, $L$ given by
 
\begin{equation}
    L = - \sum_{\ra\in\va } \log(g_{\ra, \opick})
\label{eq:loss}
\end{equation}

\noindent
where $g_{i,j}$ denotes the predicted probability of the $j^{th}$ class for the $i^{th}$ box. The proposed attack modifies the image $I$, via gradient-ascent, using the gradient, $\nabla_{I} L$. We have two variants: one attacks the most confident object (\opick is the most confident object) called \UnTargetedConf and another one attacks the most frequent object (\opick is the most frequent object) called \UnTargetedFreq. These are the most challenging setups: choosing a low-confidence or less-frequent object would make it easier to induce a misclassification. These attacks run for $max_{iter}$ iterations for a fixed $r$, and the attack is considered unsuccessful if we fail to achieve the goal. 

%Algorithm \ref{alg:non-targeted} in the appendix summarizes our non-targeted attack.

%   \begin{algorithm}[!h]
% \caption{\TargetedConf / \TargetedFreq }\label{alg:targeted}
% \DontPrintSemicolon
%   $\textbf{Input:}I_{org}, r, max_{iter}, \opick, \classK$\;
%   $\textbf{Output:}I_{adv}$\;
%   $\text{Get mask }M\text{ from }I_{org}\text{ and } \opick$\;
%   $success \gets \textbf{False}$\;
%   $I \gets I_{org}$\;
%   \For{$j\leftarrow 1$ \KwTo $max_{iter}$}
%   {
%   $\text{Compute } \va \text{ for image } I$\; 
%   \uIf {$\va = \varnothing$}
%   {
%   $\va = \argmax\limits_{\ru} g_{\ru, \classK}  \text{ where }\ru \text{ are set of }$\; 
%   $\text{predicted boxes with positive IoU with mask } M$\;
%   }
%   \uIf {$ \classK = \argmax\limits_{c} g_{\ra,c} \text{ for any } \ra \in \va \textbf{ and } \newline \opick \neq \argmax\limits_{c} g_{\ra,c} \text{ for all } \ra \in \va$}
%   {
%   $success \gets \textbf{True}$\;
%   $\textbf{break}$\;
%   }
%   $\text{Compute loss } L \text{ using }\eqref{eq:loss_targeted}$\;
%   $\text{Compute }  \nabla_{I} L \text{ using }\eqref{eq:grad_org}$ \;
%   $I \gets I - \nabla_{I}' L$\;
%   $\text{Truncate image } I \text{ in the range } [0,255]$\;
%   }
%   $I_{adv} \gets I$\;
%   \KwRet{$I_{adv}$}\;
% \end{algorithm}

\paragraph{Targeted Attack}
Our goal in the targeted attack is to generate an image $I_{adv}$ so that none of the detected boxes have the predicted class \opick and some of the boxes have the predicted class $\classK$. Here, $\classK$ denotes the targeted class for the selected object \opick. To achieve this, we use the following loss function, $L$ given by
 
\begin{equation}
    L = - \sum_{\ra\in\va } \log(g_{\ra, \classK})
\label{eq:loss_targeted}
\end{equation}

\noindent
where $g_{i,j}$ denotes the predicted probability of the $j^{th}$ class for the $i^{th}$ box. The proposed attack modifies the image $I$, via gradient-descent, using the gradient, $\nabla_{I} L$. Similar to the non-targeted attack, we have two variants: \TargetedConf and \TargetedFreq. These attacks run for $max_{iter}$ iterations for a fixed $r$, and the attack is considered unsuccessful if our goal is not achieved. 
%Algorithm \ref{alg:targeted} summarizes our targeted attack. 
During the attack, if there are no boxes with label \opick, we set $\va$ to be the box having the maximum probability of $\classK$ among all the boxes having a positive Intersection over Union (IoU) with the mask $M$. We give details of the algorithms in the supplementary material.

%Since Equations \ref{eq:loss} and \ref{eq:loss_targeted} consider the probability for $\va$, we only get gradient for the region corresponding to $\va$. As a result, only the relevant portion of the image is modified during targeted and non-targeted attacks. 

%%%%%%%%%%%%%%%%%%%%%%%%%%%%%%%%%%%%%%%%%%%%%%%%%%%%%%%%%%%%%%%%%%%%%%%%%%
%%%%%%%%%%%%%%%%%%%%%%%%%%%%%%%%%%%%%%%%%%%%%%%%%%%%%%%%%%%%%%%%%%%%%%%%%%
%%%%%%%%%%%%%%%%%%%%%%%%%%%%%%%%%%%%%%%%%%%%%%%%%%%%%%%%%%%%%%%%%%%%%%%%%%

\vspace{-3mm}
\section{Evaluation Setup}

% \paragraph{Intrinsic Evaluation} For intrinsic evaluation, we study the success of the attacks against Faster R-CNN, and the magnitude of changes to the images caused by the attacks.

% \paragraph{Downstream Evaluation} We are also interested in seeing how detectable the adversarial changes are in a downstream task: perturbations might be difficult for a human to detect in an image but can be very obvious from distortions in the downstream task.  We consider image captioning as our downstream task: captions that are completely unlike the original ones could make manipulation obvious. We use the image captioner of \cite{anderson2018bottom}, which 
% % uses an attention mechanism to attend to the bounding boxes obtained using Faster R-CNN to generate the caption, and 
% gives the state-of-the-art results. We investigate how much our \pickObjectAttack changes captions compared to an object detection attack that modifies the entire image, like that of \cite{xie2017adversarial}. 
% % We note here that our goal differs from image captioning attacks like that of \cite{chen2018attacking}.  Their goal is to force the captioner to generate specific terms, whereas we just use the image captioner to measure downstream perceptibility of object detection attacks.
% % In addition, their attack is against an earlier image captioning system that does not use object detection features

% In the supplementary material, we discuss metrics used for intrinsic and downstream evaluations and the implementation details where the values of the hyperparameters are specified.

\subsection{Intrinsic Evaluation}
%\subsection{Metrics}

\paragraph{Success Rate} 
We use success rate defined as the percentage of attacks that successfully generate adversarial examples. This is a common metric for evaluating adversarial attacks (higher means better performance).

\paragraph{ACAC and ACTC} 
We adapt these measures for object detectors from attacks against classifiers \citep{ling2019deepsec}.
For the non-targeted attacks, Average Confidence of True Class (ACTC) is calculated for object class \opick for all predicted boxes with positive IoU with the mask. This is a performance metric measuring the success of the attack methods to escape from \opick (lower means better performance). For the targeted attacks, Average Confidence of Adversarial Class (ACAC) is calculated for object class $\classK$ for all predicted boxes with label $\classK$. This shows the confidence of the attack methods to generate $\classK$ (higher means better performance).

\paragraph{Perceptibility}
To quantify the perceptibility of change in image, we follow previous work \citep{szegedy2013intriguing,xie2017adversarial} in calculating a score $\delta$ for an adversarial perturbation given by

\begin{equation}
\delta_i = 
\frac{\norm{I_{i,adv} - I_{i,org}}_2}{\sum M_i}
\label{eq:norm}
\end{equation}

\noindent
where $I_{i,adv}$ is the $i^{th}$ adversarial image, $I_{i,org}$ is the $i^{th}$ original image, and $M_i$ is the mask of the $i^{th}$ image in pixels. We normalize the $\ell_2$ norm of the image difference by the size of the mask, as our proposed attack adds noise only inside the mask and the size of the mask varies across images (lower means better performance).

\paragraph{The Structural SIMilarity (SSIM)}
We calculate the Structural SIMilarity (SSIM) to measure the similarity between the original image and the adversarial example since it is a metric which correlates well with human perception.  The definition of $\SSIM(I_{org}, I_{adv})$ between a single original image $I_{org}$ and an adversarial sample $I_{adv}$ is given in \cite{wang2004image}. We calculate the mean SSIM across all pairs of original and adversarial images
%($AVE_{SSIM}=\frac{1}{n}\sum_{i=1}^{n} SSIM(I_{i,org}, I_{i,adv})$). 
% ($\AVGSSIM=\frac{1}{n}\sum_{i=1}^{n} \SSIM(I_{i,org}, I_{i,adv})$) 
(higher means better performance).

\vspace{-2mm}
\paragraph{mAP}
Mean average precision (mAP) is calculated for objects outside the mask $M$. The high value of mAP signifies that other objects outside the mask were detected correctly. mAP is calculated using original prediction as ground truth (higher means better performance).

% \subsection{Implementation Details}
\vspace{-2mm}
\paragraph{Implementation Details}
% By adding perturbations to images, without further constraints, the top bounding box could change in each iteration in our attack method. To prevent this, we save the coordinates of the top bounding box at the beginning of our attack method. Then, in the $getIndex$ function in the Algorithm \ref{psuedo-code}, we find the index of the closest bounding box to the saved coordinates within a very small threshold ($T$). We set this threshold to 10.0. If we cannot find the closest bounding box with $T$, we terminate our attack and report it as an unsuccessful attack. We set the minimum ($r_{min}$) and maximum ($r_{max}$) of the  to 10.0 and 10000.0, respectively. Our attack method is run for a maximum iteration ($max_{iter}$) of 10 for each value of the learning rate. 
%
We test our proposed attack on a set of 1000 randomly selected images which belong to both Visual Genome and the validation set of the MSCOCO dataset. For the targeted attacks, we run attacks for 10 randomly chosen objects ($\classK$) per image resulting in 10k samples.
% The images cover a wide variety of objects, e.g., ``person'', ''table'', ``orange'', ``dog'', ``building'', etc. 
We fix the learning rate in Equation 1 ($r$) to 10k and set the maximum of iterations ($max_{iter}$) to $60$. As a comparison to our \pickObjectAttack, we design a non-targeted attack (\UnTargetedAll) against \emph{all} objects based on \cite{xie2017adversarial}. We choose a fixed label for all the boxes and do gradient descent until none of the original objects are detected. (We give the algorithm in the supplementary material.)
%(Algorithm \ref{alg:non-targeted-all}).~
We use the same learning rate as our previous attacks for a fair comparison and increase $max_{iter}$ to $120$. Since attacking all objects is a difficult task, we obtained a low success rate for \UnTargetedAll (targeted attack against all objects is not feasible) \newgreen{when $max_{iter}$ was set to $60$}.

% (we show detected attributes and objects in Figure  \ref{fig:example_eval} and \ref{fig:caption_compare}).

% In this paper, we fix the confidence threshold to $0.70$.

%%%%%%downstream evaluation
\vspace{-2mm}
\subsection{Downstream Evaluation}
% \vspace{-2mm}
%\subsection{Metrics}

\paragraph{Metrics}
The standard image captioning metrics (BLEU~\citep{papineni2002bleu}, METEOR~\citep{denkowski2014meteor}, CIDEr \citep{vedantam2015cider}, ROUGE~\citep{lin2004rouge} and SPICE~\citep{anderson2016spice}) are used to compare generated captions with human-produced reference captions, and higher scores indicate greater overlap with these reference captions.
We will use these slightly differently.  Here, we are interested in the overlap of the caption for the adversarial image and \emph{the caption for the original image}, used as the reference caption. A higher score means that the two captions are more similar, i.e. the caption for the adversarial image is less distorted. In addition, we calculate the percentage of cases for which the proposed attacks can remove the keyword corresponding to $\ro_{pick}$ from the adversarial caption when the keyword is present in the original caption (KWR).

% \begin{figure*}[!tb]
% \begin{center}
% % \fbox{\rule{0pt}{2in} \rule{.9\linewidth}{0pt}}
% \includegraphics[width=0.7\linewidth]{histo_probs.pdf}
% \end{center}
%   \caption{The histograms of number of boxes and mean probabilities for different variants of \pickObjectAttack.}
% \label{fig:histo_probs}
% \end{figure*}

% \begin{figure*}[!tb]
% \begin{center}
% % \fbox{\rule{0pt}{2in} \rule{.9\linewidth}{0pt}}
% \includegraphics[width=0.7\linewidth]{histo_iteration.pdf}
% \end{center}
%   \caption{The histogram of number of iterations for different variants of \pickObjectAttack.}
% \label{fig:histo_iteration}
% \end{figure*}

%\subsection{Implementation Details}
\vspace{-2mm}
\paragraph{Implementation Details}
We generate captions using three non-targeted attacks: \UnTargetedAll, \UnTargetedFreq, \UnTargetedConf and two targeted attacks: \TargetedFreq, \TargetedConf. To do so, we use $100$ successful adversarial examples for the non-targeted attacks for a shared set having $100$ images \newgreen{(these images belong to MSCOCO dataset)}. We use $1000$ successful adversarial examples for the targeted attacks for the shared set ($10$ per image).

% \vspace{-5mm}
\section{Results}
% \vspace{-2mm}

\subsection{Intrinsic Evaluation}
\label{sec:res-intrinsic}

% \begin{table}[htbp]
% \caption{The success rate, ACAC and ACTC for different proposed attacks.}
% \label{tab:results_success}
% \begin{center}
% \begin{tabular}{|c|c|c|c|c|}
% \hline
% \multicolumn{1}{|c|}{\bf APPROACHES}  &\multicolumn{1}{|c|}{\bf SUCCESS RATE} &\multicolumn{1}{|c|}{\bf ACAC} &\multicolumn{1}{|c|}{\bf ACTC}
% \\ \hline
% % \PermTargetedConf &  & &   & \_   \\ 
% % \PermTargetedFreq &  & &  &  \_   \\
% % \PermUnTargetedConf & &  &  \_  &   \\ 
% % \PermUnTargetedFreq &  & & \_  &   \\
% \TargetedFreq & 89.90\%  & 26.55\% &  \_   \\
% \hline
% \TargetedConf & 76.97\%  & 24.53\%  & \_   \\ 
% \hline
% \UnTargetedFreq & 95.30\%  & \_  & 1.25\%  \\
% \hline
% \UnTargetedConf & 98.40\%  &  \_  & 2.59\%  \\ 
% \hline
% \end{tabular}
% \end{center}
% \end{table}

\begin{table}[!tb]
\begin{center}
\caption{Success Rate (SR), ACAC, ACTC and mAP for different proposed attacks.
}\label{tab:results_success}
\resizebox{0.35\textwidth}{!}{
\begin{tabular}{|l|c|c|c|c|}
\hline
\textbf{APPROACHES}  & \textbf{SR} & \textbf{ACAC} & \textbf{ACTC} & \textbf{mAP}
\\ \hline 
% \hline
% \PermTargetedConf &  & &   & \_   \\ 
% \PermTargetedFreq &  & &  &  \_   \\
% \PermUnTargetedConf & &  &  \_  &   \\ 
% \PermUnTargetedFreq &  & & \_  &   \\
\UnTargetedAll & 79.80\% & \_ & \_ & 0.30\% \\
\hline
\TargetedFreq & 89.90\%  & 26.55\% &  \_ & 86.09\%  \\
\hline
\TargetedConf & 76.97\%  & 24.53\%  & \_  & 91.97\%  \\ 
\hline
\UnTargetedFreq & 95.30\%  & \_  & 1.25\% & 94.20\% \\
\hline
\UnTargetedConf & 98.40\%  &  \_  & 2.59\% & 95.47\%  \\ 
\hline
\end{tabular}
}
\end{center}
\vspace{-5mm}
\end{table}

\paragraph{Quantitative Results}
Table~\ref{tab:results_success} shows the success rate, ACAC and ACTC for \newgreen{\UnTargetedAll and} our variants of the \pickObjectAttack. \newgreen{The goal of the \UnTargetedAll algorithm is none of the original objects are detected. This leads to a low success rate.} Comparing the variants of the \pickObjectAttack, the non-targeted attacks are more successful compared to the targeted ones. Since we only need to induce a misclassification for the non-targeted attacks, we can achieve a better success rate. \TargetedConf has the lowest success rate. For \TargetedConf, out of $2303$ unsuccessful attacks, $1750$ attacks are unsuccessful only because \TargetedConf cannot find any bounding box with a positive IoU with the mask. This never happens for \TargetedFreq since the mask is larger for the most frequent object in comparison with the most confident one in the image. %If we exclude these unsuccessful cases for \TargetedConf, the success rate will be $93.30\%$. 
Out of the cases where there is a bounding box with positive IoU with the mask for \TargetedConf, the success rate is $93.30\%$. 
\UnTargetedConf produces the highest success rate since it does not face this condition. The ACAC and ACTC metrics here show that the attack approaches can generate high confidence for adversarial and low confidence for original classes. \newgreen{Note that we do not report ACAC and ACTC for \UnTargetedAll since there is no notion of adversarial and true class when attacking all the objects.} To understand the way in which the perturbations contribute to attack success, following \cite{xie2017adversarial} we also randomly permute the perturbations generated by the proposed attacks for the adversarial images. This leads to near zero success rates for all attacks, showing that the spatial structure of the perturbations plays a major role in fooling Faster R-CNN rather than the magnitude of the perturbations.

Table~\ref{tab:results_success} also shows the mAP metric for \newgreen{\UnTargetedAll and} our proposed attacks. \newgreen{\UnTargetedAll adds perturbations to the entire image. Hence, for \UnTargetedAll, we calculate mAP for all detected objects. Table~\ref{tab:results_success} shows that \UnTargetedAll generates very low mAP value owing to its high success rate.} On the other hand, the variants of \pickObjectAttack add perturbations only inside the mask with the purpose of preserving the labels of the bounding boxes outside the mask. This perturbation may lead to a different set of bounding boxes by the region proposal network (RPN). These bounding boxes are more likely to have a positive IoU with the mask. Here, mAP shows the impact of perturbation on the bounding boxes outside the mask. As shown in Table~\ref{tab:results_success}, the proposed attacks mostly do not change the bounding boxes since they generate high mAP values. These results demonstrate that there are two factors impacting on the mAP: the amount of perturbations and the size of the mask. Our targeted attacks add more perturbations to images in order to fool Faster R-CNN to detect targeted classes, and they have lower values for the mAP in comparison with the non-targeted attacks. The attacks against the most frequent objects (\TargetedFreq and \UnTargetedFreq) also generate lower mAP than the most confident objects (\TargetedConf and \UnTargetedConf) since the size of the mask for the frequent objects is larger than the confident objects.

\begin{table}[!tb]
\begin{center}
\caption{Mean and standard deviation of $\ell_2$-norm of the difference image normalised by the image size, and SSIM between original and adversarial images. 
% , which denotes the $\ell_{2}$-norm of the difference image (i.e. $I_{adv}-I_{org}$) normalised by size of the mask.
}\label{tab:results_perturbation_ssim}
\resizebox{0.35\textwidth}{!}{
\begin{tabular}{|l|c|c|c|}
\hline
\multirow{2}{*}{\bf APPROACHES}  & \multicolumn{2}{c|}{$\ell_2$-norm} & \multirow{2}{*}{\bf SSIM} \\ \cline{2-3}
& \textbf{MEAN} & \textbf{STD. DEV.} &
\\ \hline
% \hline
\UnTargetedAll  & $1.39 \times 10^{-3}$  &  $5.28 \times 10^{-4}$ & 94.33\%
\\ 
\hline
% 0.0004195085784159495
% 0.0011838444846366317
\TargetedFreq &  $1.18 \times 10^{-3}$ & $ 4.20 \times 10^{-4}$  & 98.53\%
\\
\hline
% 0.0004188206838901302
% 0.0011422325271178472
\TargetedConf & $1.14 \times 10^{-3}$ & $ 4.19 \times 10^{-4}$  & 98.73\%
\\
\hline
% 0.00023230152432004511
% 0.0004925571741603213
\UnTargetedFreq  & $4.93 \times 10^{-4}$ & $ 2.32 \times 10^{-4}$  & 99.22\%
\\ 
\hline
% 0.0002291263480328466
% 0.00040743036479032064
\UnTargetedConf  &  $4.07 \times 10^{-4}$ & $ 2.29 \times 10^{-4}$  & 99.32\%
\\ 
\hline

\end{tabular}
}
\end{center}
\vspace{-5mm}
\end{table}

\begin{table}[!tb]
\begin{center}
\caption{Mean and standard deviation, for the successful cases, of $\delta$. 
%and SSIM between original and adversarial images. 
% , which denotes the $\ell_{2}$-norm of the difference image (i.e. $I_{adv}-I_{org}$) normalised by size of the mask. 
}\label{tab:results_perturbation}
\resizebox{0.32\textwidth}{!}{
\begin{tabular}{|l|c|c|}
\hline
\multirow{2}{*}{\bf APPROACHES}  & \multicolumn{2}{c|}{$\delta$} \\ \cline{2-3}
  & \textbf{MEAN} & \textbf{STD. DEV.}  \\
 \hline
%  \hline
% \PermTargetedConf  & $1.06 \times 10^{-2}$  & $ 2.82 \times 10^{-2}$ & 98.04\%\\ % 0.010634913585  0.0282331164652   , 0.980384205048
% \PermTargetedFreq  & $1.53 \times 10^{-3}$ & $ 1.41 \times 10^{-3}$ & 97.18\%  \\ %0.00152541422781 %0.00140859296218    ,  0.971847542921
% \PermUnTargetedConf  & $6.65 \times 10^{-3}$ & $ 1.58 \times 10^{-2}$ & 98.92\%    \\ % 0.0158165974249 0.00665405171952 , 0.989168536585
% \PermUnTargetedFreq  & $6.62 \times 10^{-4}$ & $ 8.97 \times 10^{-4}$ & 98.49\%  \\ %0.000897409410566
% % 0.000662291512132, 0.984877239979
\TargetedFreq  & $1.53 \times 10^{-3}$ & $ 1.41 \times 10^{-3}$    \\ %0.00152541422781 %0.00140859296218    ,  0.985334835434
\hline

\TargetedConf  & $1.06 \times 10^{-2}$  & $ 2.82 \times 10^{-2}$  \\ % 0.010634913585  0.0282331164652   , 0.987292565436
\hline

\UnTargetedFreq  & $6.62 \times 10^{-4}$ & \bf $8.97 \times 10^{-4}$  \\ % 0.000662214217707  0.000897390687383  , 0.992206962644
\hline

\UnTargetedConf  & $6.65 \times 10^{-3}$ & $ 1.58 \times 10^{-2}$    \\ % 0.00665375730819 , 0.0158164681921 , 0.993164136179
\hline
\end{tabular}
}
\end{center}
\vspace{-10mm}
\end{table}

As shown in Table~\ref{tab:results_perturbation_ssim}, SSIM is high for all attack approaches. This shows that the approaches are successful in adding imperceptible perturbations to images. \newgreen{Although, compared to \UnTargetedAll, our proposed approaches achieve higher SSIM.} (We show histograms of the number of boxes/iterations and visualize the perturbations of the adversarial examples in the supplementary material). \newgreen{To compare against \UnTargetedAll, we normalize $\ell_2$-norm by the image size in Table~\ref{tab:results_perturbation_ssim}. As per the results, the proposed approaches lead to lower $\ell_2$-norm than \UnTargetedAll.} 

Table~\ref{tab:results_perturbation} shows the mean and standard deviation of $\delta$ for \emph{successful} cases. \TargetedConf generates the highest $\delta$. Similarly, \UnTargetedConf has higher $\delta$ in comparison with \UnTargetedFreq. This means that attacking the most confident object is harder than attacking the most frequent object in the image, even though there are typically more instances of the most frequent object. In fact, from Table~\ref{tab:results_perturbation}, we can see that \UnTargetedConf requires more noise than \TargetedFreq. In the supplementary material, we also show the robustness of the adversarial images against resizing with different scales.

\paragraph{Qualitative Results}
Consider the pair of images in the upper row of Figure~\ref{fig:caption_compare}.
For generating the adversarial image on the right, we targeted ``cat'' for ``sheep'' in this example. The outputs of Faster R-CNN (the labels of bounding boxes) show that ``sheep'' is changed to ``cat''. 
Similarly, in the lower row, the targeted attack approach successfully changes all instances of ``bird'' to ``sign'' as shown in the labels of bounding boxes.
\green{Note that Figure~\ref{fig:caption_compare} shows only bounding boxes which are fed to the captioning system. Since the captioning system picks the top scoring bounding boxes from Faster R-CNN, we obtain different bounding boxes for the original and the adversarial images. }

\vspace{-5mm}
\subsection{Downstream Evaluation}

\begin{figure*}[t]
\begin{center}
% \fbox{\rule{0pt}{2in} \rule{.9\linewidth}{0pt}}
\includegraphics[width=0.68\linewidth]{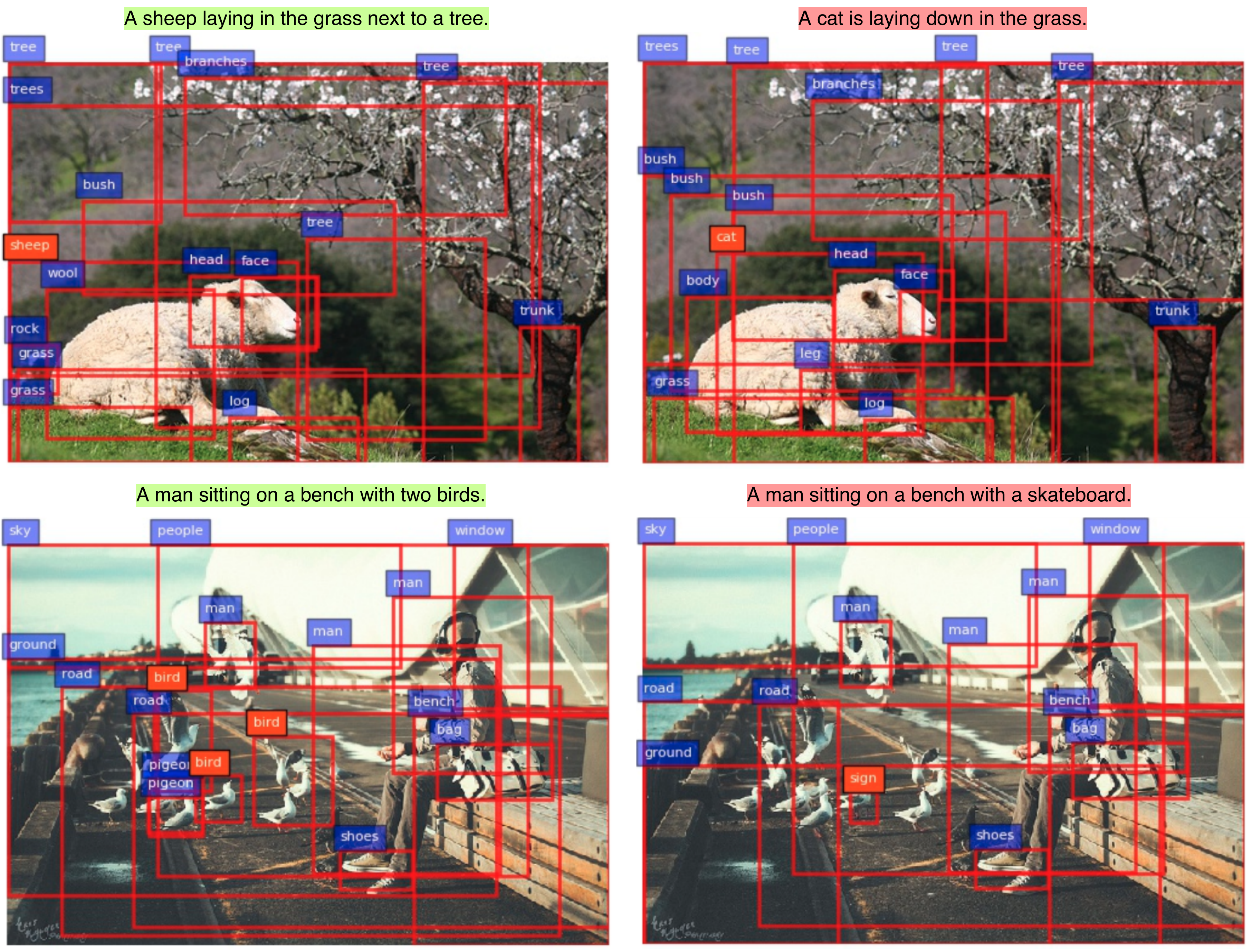}
\end{center}
  \vspace{-5mm}
  \caption{The first column includes the original images and the second column includes the adversarial images with their corresponding generated captions. The bounding boxes and the labels on these images are the outputs of Faster R-CNN.}
\label{fig:caption_compare}
\vspace{-5mm}
\end{figure*}

% \paragraph{Image Captioning} 
\paragraph{Quantitative Results}
% Table \ref{tab:results_image_caption_metrics} shows that fewer changes in the image lead to fewer changes in their corresponding captions. Since \UnTargetedAll changes the whole image, it generates the lowest values for the image captioning metrics. The differences are quite dramatic: BLEU-1 corresponds to unigram, and is much smaller for \UnTargetedAll than for either of the \pickObjectAttack; BLEU-3, which involves trigram, shows that there are zero overlaps of trigrams between perturbed and original captions.
%

\begin{table*}[!tb]
\begin{center}
\caption{Image captioning metrics and KWR (in $\%$) for different attacks (B-N is  BLEU-N).}\label{tab:results_image_caption_metrics}
\resizebox{0.7\textwidth}{!}{
\begin{tabular}{|l|c|c|c|c|c|c|c|c|c|}
\hline
\textbf{APPROACHES}  & \textbf{B-1} & \textbf{B-2} & \textbf{B-3} & \textbf{B-4} & \textbf{CIDEr} & \textbf{METEOR} & \textbf{ROUGE-L} & \textbf{SPICE} & \textbf{KWR}
\\ \hline 
% \hline
\UnTargetedAll & 23.15 & 6.91 & 0.00 & 0.00 & 5.59 & 8.19 & 22.70  & 0.86 & \_  \\
% \TargetedConf &  &    & &&  &   &    \\ 
% \TargetedFreq &  &    & & &  &  &     \\
\hline
\TargetedFreq & 44.77 & 31.82 & 24.45 & 19.58 & 179.26 & 20.67 & 44.03 & 22.98 & 72.43
\\
\hline
\TargetedConf & 57.73 & 47.57 & 40.92 & 35.81 & 331.74 & 30.30 & 57.28 & 40.19 & 80.17
\\
\hline
\UnTargetedFreq & 63.28 & 53.23 & 46.27 & 40.62 & 389.55 & 33.06 & 62.91 & 48.16 & 54.00  \\
\hline
\UnTargetedConf & 70.39 &  62.59  & 56.98 & 52.48 & 495.17 & 38.03 & 69.39 & 57.18 & 76.00 \\ 
\hline
\end{tabular}}
\end{center}
\vspace{-6mm}
\end{table*}

Table \ref{tab:results_image_caption_metrics} shows the image captioning metrics for different attack approaches. Since \UnTargetedAll changes the whole image, it generates the lowest values for the metrics. The differences are quite dramatic: BLEU-1 is much smaller for \UnTargetedAll than for any variant of \pickObjectAttack; BLEU-3 is zero for \UnTargetedAll which shows that there are zero overlaps of trigrams (sequences of three words) between perturbed and original captions. Comparing our \pickObjectAttack variants, \TargetedFreq and \UnTargetedFreq change more regions in the image because they attack the most frequent object. Thus, they generate lower values in comparison with \TargetedConf and \UnTargetedConf, respectively. The targeted attacks have lower values in comparison with the non-targeted ones since they add more perturbations to images to generate particular classes. From these results, it is evident that fewer changes in the image lead to fewer changes in the corresponding captions. %This shows that the amount of perturbation has more impacts than the size of regions on the generated captions.

In terms of keyword removal (KWR), 
\TargetedConf and \UnTargetedConf have higher values in comparison with \TargetedFreq and \UnTargetedFreq since they add more perturbations to change the label of the most confident object in the image. \TargetedFreq and \TargetedConf have higher values than their non-targeted versions since they aim to generate a particular class (since $\ro_{pick}$ is not fixed for \UnTargetedAll, we do not provide KWR for this approach). To study perceptibility of attack, we calculate mean, standard deviation of $\ell_2$-norm of the difference image and SSIM between the adversarial images, used for the downstream evaluation, and the original images.
Since \UnTargetedAll modifies the whole image, to compare across attacks, we normalize the $\ell_2$-norm of the difference image by the image size for all attacks (we include $\ell_2$-norm normalised by mask size, as per Equation~\ref{eq:norm}, in Table \ref{tab:results_perturbation_caption_subset}, for direct comparison with Table~\ref{tab:results_perturbation}).
As shown in Table~\ref{tab:results_perturbation_caption_subset_whole}, all methods generate perturbations with low perceptibility according to the standard metrics for adversarial examples.  The non-targeted variants of \pickObjectAttack are less detectable than the targeted ones; \UnTargetedAll is more similar to the targeted variants of \pickObjectAttack, although the perturbations are still small.  The perceptibility of \UnTargetedAll by these standard metrics, however, contrasts strongly with the effects on the downstream image captioning task that we describe above, strongly suggesting that the evaluation of how detectable adversarial perturbations are should extend beyond the standard perceptibility metrics.

\begin{table}[!tb]
\begin{center}
\caption{Mean and standard deviation of $\ell_2$-norm of the difference image normalised by the image size, and SSIM between original and adversarial images for the captioning examples. 
% , which denotes the $\ell_{2}$-norm of the difference image (i.e. $I_{adv}-I_{org}$) normalised by size of the mask.
}\label{tab:results_perturbation_caption_subset_whole}
\resizebox{0.35\textwidth}{!}{
\begin{tabular}{|l|c|c|c|}
\hline
\multirow{2}{*}{\bf APPROACHES}  & \multicolumn{2}{c|}{$\ell_2$-norm} & \multirow{2}{*}{\bf SSIM} \\ \cline{2-3}
& \textbf{MEAN} & \textbf{STD. DEV.} &
\\ \hline
% \hline
\UnTargetedAll  & $1.40 \times 10^{-3}$ & $ 3.97 \times 10^{-4}$    & 98.23\%
\\ 
\hline
\TargetedFreq & $1.22 \times 10^{-3}$ & $4.86 \times 10^{-4}$ & 98.16\%
\\
\hline
\TargetedConf & $ 1.20 \times 10^{-3}$ & $4.61 \times 10^{-4}$ & 98.42\%
\\
\hline
\UnTargetedFreq  & $4.55 \times 10^{-4}$  & $ 2.24 \times 10^{-4}$ & 99.12\%
\\ 
\hline
\UnTargetedConf  & $4.08 \times 10^{-4}$ & \bf $2.50 \times 10^{-4}$ & 99.23\%
\\ 
\hline

\end{tabular}
}
\end{center}
\vspace{-5mm}
\end{table}

\begin{table}
\begin{center}
\caption{Mean and standard deviation of $\delta$ for the captioning examples.
% , which denotes the $\ell_{2}$-norm of the difference image (i.e. $I_{adv}-I_{org}$) normalised by size of the mask.
}\label{tab:results_perturbation_caption_subset}
\resizebox{0.3\textwidth}{!}{
\begin{tabular}{|l|c|c|}
\hline
\multirow{2}{*}{\bf APPROACHES}  & \multicolumn{2}{c|}{$\delta$} \\ \cline{2-3}
& \textbf{MEAN} & \textbf{STD. DEV.}
\\ \hline 
% \hline
% \UnTargetedAll  & $1.40 \times 10^{-3}$ & $ 3.97 \times 10^{-4}$ & 98.23\%   
% \\ 
\TargetedFreq & $1.49 \times 10^{-3}$ & $8.40 \times 10^{-4}$ 
\\
\hline
\TargetedConf & $ 8.37 \times 10^{-3}$ & $1.85 \times 10^{-2}$ 
\\
\hline
\UnTargetedFreq  & $5.52 \times 10^{-4}$  & $ 3.35 \times 10^{-4}$
\\ 
\hline
\UnTargetedConf  & $4.18 \times 10^{-3}$ & \bf $8.39 \times 10^{-3}$ \\ % 
\hline
\end{tabular}
}
\end{center}
\vspace{-10mm}
\end{table}

\vspace{-2mm}
\paragraph{Qualitative Results}

Figure  \ref{fig:caption_compare} shows two examples fed into the captioning model (the attention weights of the model for these examples are visualized in the Figures 5 and 6 in the supplementary material). The original image in the first row leads to the caption of ``a sheep laying in the grass next to a tree''. As discussed in \S\ref{sec:res-intrinsic}, a targeted attack changes ``sheep'' to ``cat''; the caption is correspondingly changed to ``a cat is laying down in the grass''. This means that our attack against Faster R-CNN can indirectly attack the captioning model to generate a different caption with our targeted class (``cat''). This is also true for the image in the second row for generating a different caption. The original caption for the image is ``a man sitting on a bench with two birds''. As noted in \S\ref{sec:res-intrinsic}, the attack approach successfully changes ``bird'' to ``sign''; the caption for the adversarial example here is ``a man sitting on a bench with a skateboard'' which is different from the original one. Although the attack model leads to a new caption, the caption does not include our targeted class (``sign''); ``skateboard'' is chosen because it is strongly favoured by the language model. As indicated by Table~\ref{tab:results_image_caption_metrics}, \UnTargetedAll changes captions much more dramatically. (In the supplementary material, we show some more examples of predicted captions.)

\vspace{-3mm}
\section{Conclusion and Future work}
\vspace{-3mm}

We have proposed \pickObjectAttack, a type-specific adversarial attack for Faster R-CNN. The proposed approach attacks a specific object in an image and aims to preserve the labels of other detected objects in the image. We study both targeted and non-targeted attacks. For each one, we have two variants: attacking the most frequent and the most confident object in the image. Amongst them, the lowest success rate is obtained by the \TargetedConf because this approach sometimes fails to find bounding boxes within the mask. The results show that attacking the most confident object requires more noise than the most frequent object. 
%In fact, we find that \UnTargetedConf requires more noise than \TargetedFreq. 
The proposed attacks achieve high mAP values for bounding boxes outside the mask which shows that they preserve the labels of other detected objects. \newgreen{The results also show that the variants of \pickObjectAttack lead to less noise in comparison with a baseline attack (\UnTargetedAll) adapted from \cite{xie2017adversarial} that modifies the entire image.} In addition to standard perceptibility metrics, we carried out an evaluation on a downstream task, studying the impact of the adversarial images on a state-of-the-art image captioning system. We compared the captions generated by different variants of \pickObjectAttack with \UnTargetedAll. The results show that although all models produce perturbations with low perceptibility, \UnTargetedAll produces dramatically distorted captions, in contrast with \pickObjectAttack, suggesting that evaluation on downstream tasks would be a useful complement to standard perceptibility measures.  In the context of applications like forensics, for example, this indicates that type-specific adversarial attacks against object detection are harder to detect than broader sorts of attacks.

% As future work, we plan to explore the impact of our attacks against other downstream tasks such as visual question answering (VQA). In this paper, we only attack the detected objects by Faster R-CNN. We aim to study attacking both detected objects and attributes by Faster R-CNN in both targeted and non-targeted scenarios. This might be a difficult scenario since it is relatively straightforward for an object detector to learn a set of attributes corresponding to a specific object. 
In future work, we aim to study the more challenging task of attacking attributes as well as objects detected by Faster R-CNN, simultaneously: the challenge here is that it is relatively straightforward for an object detector to learn a set of attributes corresponding to a specific object, and so multiple sources of information have to be perturbed.  

% More broadly, we plan to explore the impact of these and other attacks against other downstream tasks such as visual question answering (VQA), as part of understanding the extent to which images are vulnerable to manipulation and the perceptibility of this to human viewers of those images.

\vspace{-3mm}
% \bibliographystyle{model2-names}
% \bibliography{elsarticle-template}

\newpage

\section*{Supplementary Material}
% Supplementary material that may be helpful in the review process should
% be prepared and provided as a separate electronic file. That file can
% then be transformed into PDF format and submitted along with the
% manuscript and graphic files to the appropriate editorial office.

\begin{algorithm}[!h]
\caption{\UnTargetedConf / \UnTargetedFreq}\label{alg:non-targeted}
\DontPrintSemicolon
  $\textbf{Input:}I_{org}, r, max_{iter}, \opick$\;
  $\textbf{Output:}I_{adv}$\;
  $\text{Get mask }M\text{ from }I_{org}\text{ and } \opick$\;
  $success \gets \textbf{False}$\;
  $I \gets I_{org}$\;
  \For{$j\leftarrow 1$ \KwTo $max_{iter}$}
  {
  $\text{Compute } \va \text{ for image } I$\; 
  \uIf { $ \va = \varnothing$} %$\ro_{pick} \neq \argmax\limits_{c} g_{\ra,c} \text{ for any } \ra \in \va$}
  {
  $success \gets \textbf{True}$\;
  $\textbf{break}$\;
  }
  $\text{Compute loss } L \text{ using Equation 3} $\;
  $\text{Compute }  \nabla_{I} L \text{ using Equation 2}$ \;
  $I \gets I + \nabla_{I} L$\;
  $\text{Truncate image } I \text{ in the range } [0,255]$\; 
  }
  $I_{adv} \gets I$\;
  \KwRet{$I_{adv}$}\;
\end{algorithm}

\begin{algorithm}[!h]
\caption{\TargetedConf / \TargetedFreq }\label{alg:targeted}
\DontPrintSemicolon
  $\textbf{Input:}I_{org}, r, max_{iter}, \opick, \classK$\;
  $\textbf{Output:}I_{adv}$\;
  $\text{Get mask }M\text{ from }I_{org}\text{ and } \opick$\;
  $success \gets \textbf{False}$\;
  $I \gets I_{org}$\;
  \For{$j\leftarrow 1$ \KwTo $max_{iter}$}
  {
  $\text{Compute } \va \text{ for image } I$\; 
  \uIf {$\va = \varnothing$}
  {
  $\va = \argmax\limits_{\ru} g_{\ru, \classK}  \text{ where }\ru \text{ are set of }$\; 
  $\text{predicted boxes with positive IoU with mask } M$\;
  }
  \uIf {$ \classK = \argmax\limits_{c} g_{\ra,c} \text{ for any } \ra \in \va \textbf{ and } \newline \opick \neq \argmax\limits_{c} g_{\ra,c} \text{ for all } \ra \in \va$}
  {
  $success \gets \textbf{True}$\;
  $\textbf{break}$\;
  }
  $\text{Compute loss } L \text{ using Equation 4}$\;
  $\text{Compute }  \nabla_{I} L \text{ using Equation 2}$ \;
  $I \gets I - \nabla_{I}' L$\;
  $\text{Truncate image } I \text{ in the range } [0,255]$\;
  }
  $I_{adv} \gets I$\;
  \KwRet{$I_{adv}$}\;
\end{algorithm}

\begin{algorithm}[h]
\caption{\UnTargetedAll}\label{alg:non-targeted-all}
\DontPrintSemicolon
  $\textbf{Input:}I_{org}, r, max_{iter}$\;
  $\textbf{Output:}I_{adv}$\;
  $ \vc_{org} \gets \text{set of predicted classes for } I_{org} $\;
  $\text{Randomly select class } \rz \notin \vc_{org}$\;
  $success \gets \textbf{False}$\;
  $I \gets I_{org}$\;
  \For{$j\leftarrow 1$ \KwTo $max_{iter}$}
  {
  \uIf {$ \argmax\limits_{c} g_{\rb,c} \notin \vc_{org} \text{ for all boxes } \rb $}
  {
  $success \gets \textbf{True}$\;
  $\textbf{break}$\;
  }
  $ L \gets - \sum_{\rb} log(g_{\rb, \rz}) $\;
  $\nabla_{I} L \gets \text{rescale}(\nabla_{I'}' L,s)$\;
  $I \gets I - \nabla_{I} L$\;
  $\text{Truncate image } I \text{ in the range } [0,255]$\;
  }
  $I_{adv} \gets I$\;
  \KwRet{$I_{adv}$}\;
\end{algorithm}

\begin{figure*}[!tb]
\begin{center}
% \fbox{\rule{0pt}{2in} \rule{.9\linewidth}{0pt}}
\includegraphics[width=0.7\linewidth]{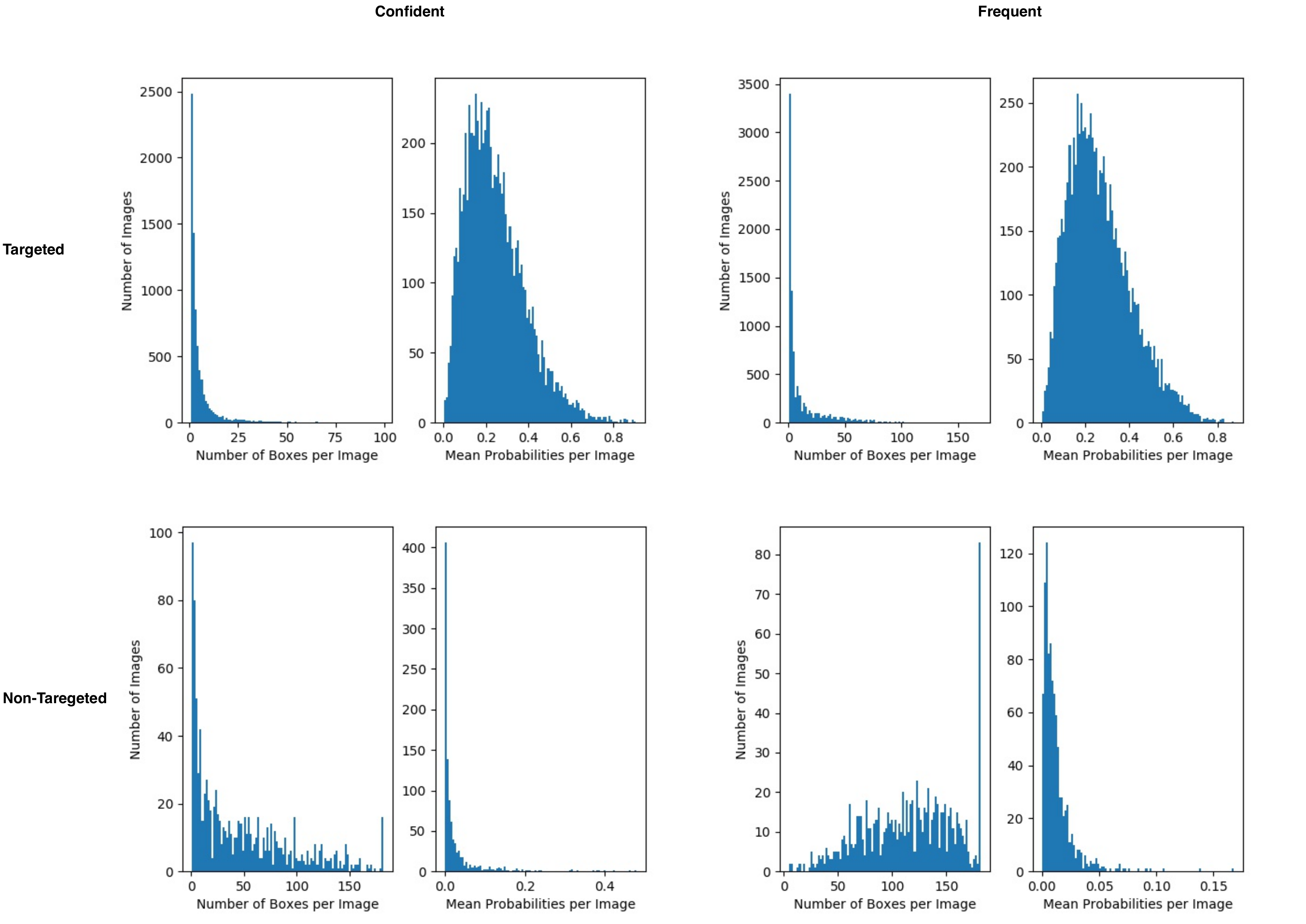}
\end{center}
  \caption{The histograms of number of boxes and mean probabilities for different variants of \pickObjectAttack.}
\label{fig:histo_probs}
\end{figure*}

\begin{figure*}[!tb]
\begin{center}
% \fbox{\rule{0pt}{2in} \rule{.9\linewidth}{0pt}}
\includegraphics[width=0.7\linewidth]{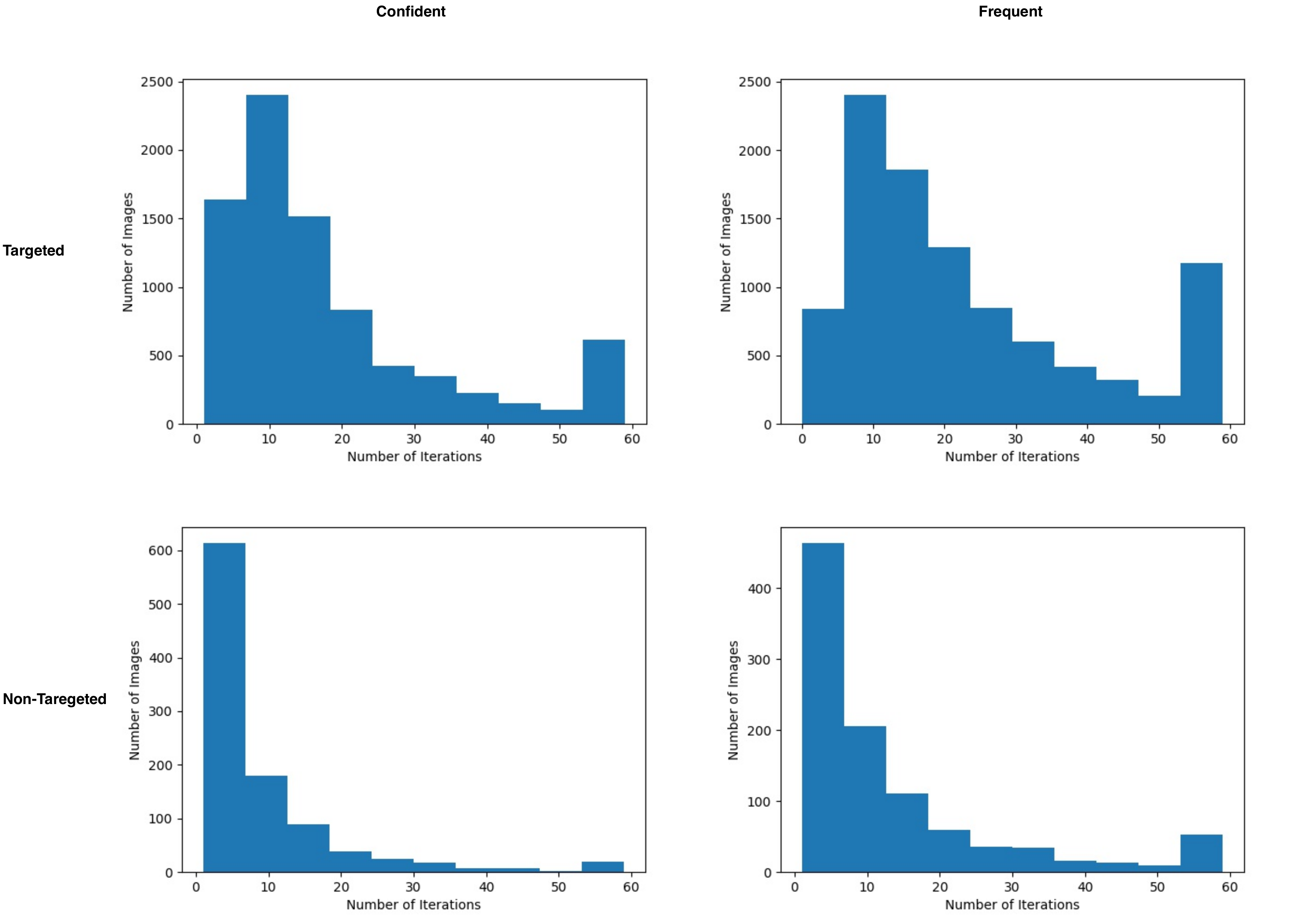}
\end{center}
  \caption{The histogram of number of iterations for different variants of \pickObjectAttack.}
\label{fig:histo_iteration}
\end{figure*}

% \subsection{Implementation Details}

% As a comparison to our \pickObjectAttack, we design a non-targeted attack against \emph{all} objects based on \cite{xie2017adversarial}. We choose a fixed label for all the boxes and do gradient descent until none of the original objects are detected (Algorithm \ref{alg:non-targeted-all}).~We name this attack \UnTargetedAll. We use the same learning rate as our previous attacks for a fair comparison and increase $max_{iter}$ to $120$. Since attacking all objects is a difficult task, we obtained a low success rate for \UnTargetedAll (targeted attack against all objects is not feasible). We generate captions using three non-targeted attacks: \UnTargetedAll, \UnTargetedFreq, \UnTargetedConf and two targeted attacks: \TargetedFreq, \TargetedConf. To do so, we use $100$ successful adversarial examples for the non-targeted attacks for a shared set having $100$ images. We use $1000$ successful adversarial examples for the targeted attacks for the shared set ($10$ per image).

%%%%% scale results for all, 0.6, 0.8, 1.2, 1.4
\begin{table}[!tb]
\begin{center}
\caption{Success Rate for our proposed attacks after resizing the adversarial images with different scales: 0.6, 0.8, 1.2 and 1.4. }\label{tab:results_success_scale}
\resizebox{0.45\textwidth}{!}{
\begin{tabular}{|l|c|c|c|c|}
\hline
\multirow{2}{*}{\bf APPROACHES}  & \multicolumn{4}{c|}{\textit{Scale}} \\ \cline{2-5}
& \textbf{0.6}  & \textbf{0.8}  & \textbf{1.2} & \textbf{1.4}
\\ \hline 
% \hline
\TargetedFreq &  16.75\%  & 44.70\% & 72.35\% & 78.30\% \\ %& 53.03\%  \\
\hline
\TargetedConf & 11.30\% & 38.08\% & 65.72\% & 74.32\% \\ % & 47.36\% \\
\hline  
\UnTargetedFreq  & 2.31\% & 9.76\% & 26.76\% & 34.63\% \\ % 18.47\% \\
\hline 
\UnTargetedConf  & 14.23\% & 26.42\% & 42.78\% & 52.34\% \\ % 34.04\% \\ 
\hline 
\end{tabular}
}
\end{center}
\end{table}

\begin{figure*}[!tb]
\begin{center}
% \fbox{\rule{0pt}{2in} \rule{.9\linewidth}{0pt}}
\includegraphics[width=0.5 \linewidth]{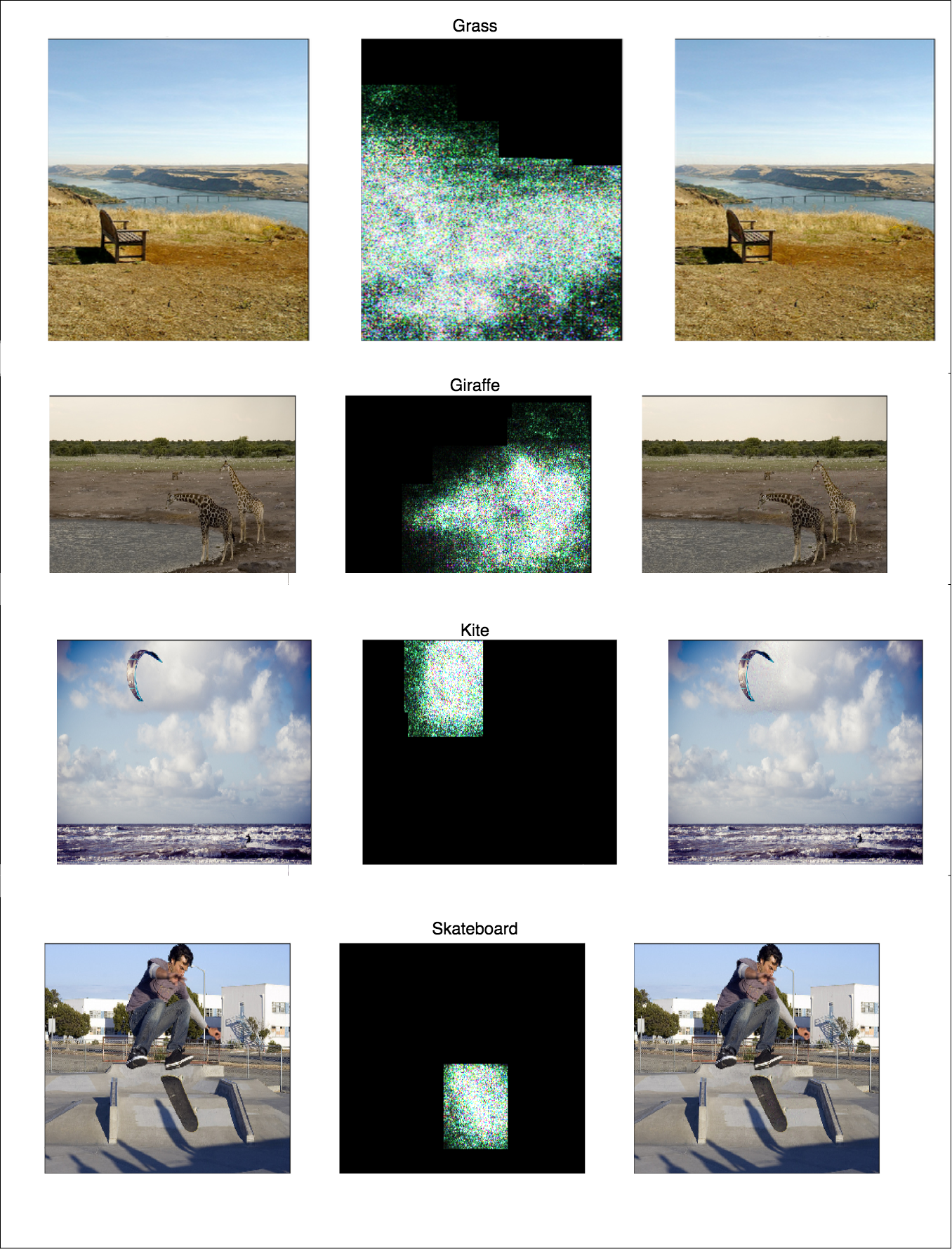}
\end{center}
  \caption{Each row contains the original image, adversarial perturbation and the adversarial image (from left to right). The first two rows are from \TargetedFreq and the last two rows are from \TargetedConf. We show $o_{pick}$ on top of each row. ``grass'' changes to ``star'' and ``giraffe'' changes to ``lion'' in the first and second rows, respectively. ``kite'' changes to ``eyes'' and ``skateboard'' changes to ``column'' for the third and fourth rows, respectively.}
\label{fig:pert_targeted}
\end{figure*}

\begin{figure*}[!tb]
\begin{center}
% \fbox{\rule{0pt}{2in} \rule{.9\linewidth}{0pt}}
\includegraphics[width=0.7 \linewidth]{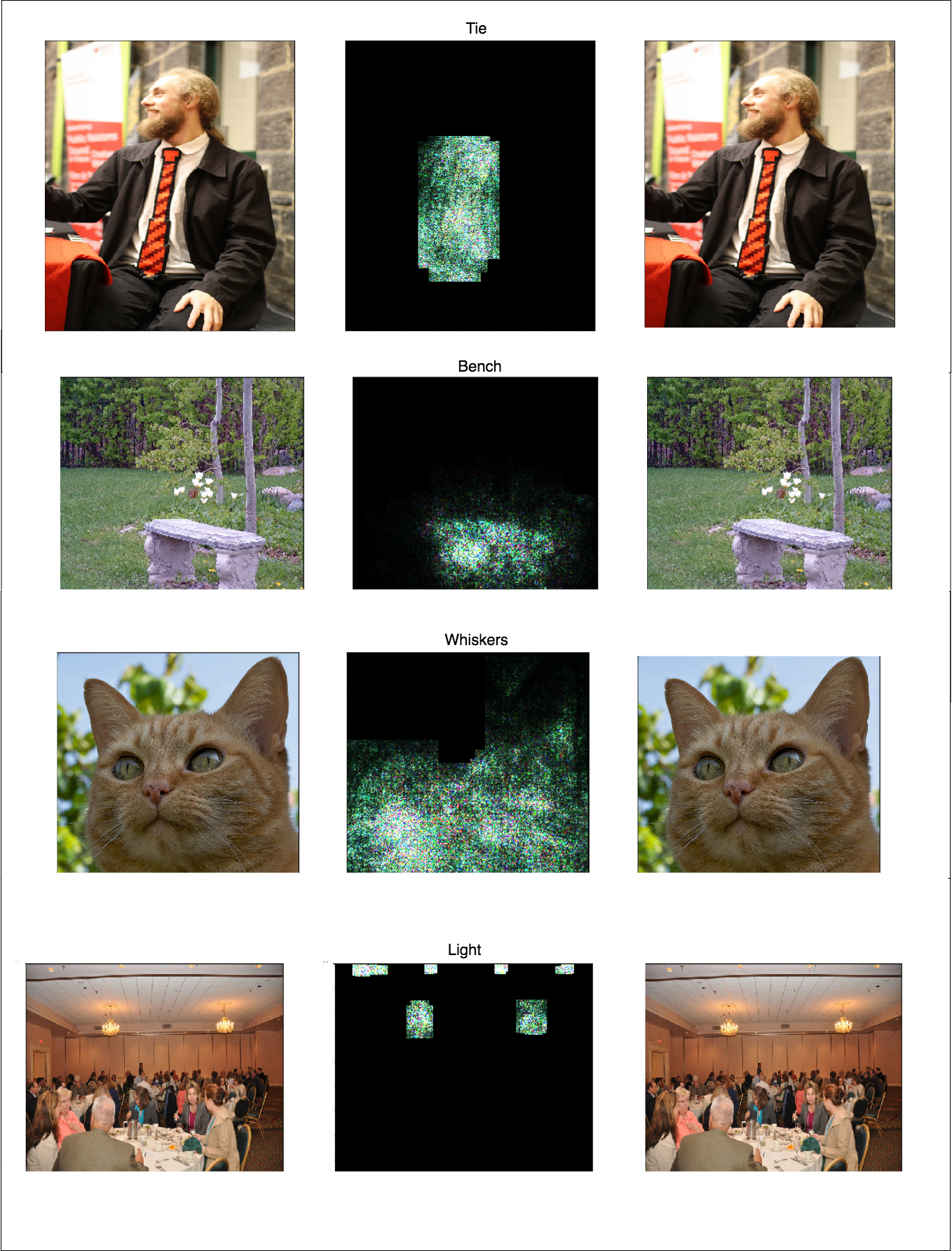}
\end{center}
  \caption{Each row contains the original image, adversarial perturbation and the adversarial image (from left to right). The first two rows are from \UnTargetedFreq and the last two rows are from \UnTargetedConf. We show $o_{pick}$ on top of each row. For example, ``tie'' is $o_{pick}$ in the first row.}
\label{fig:pert_untargeted}
\end{figure*}

\begin{table*}[!h]
\begin{center}
\caption{Examples of predicted captions for adversarial and original image using different attack approaches. }
\label{tab:results_generated_captions}
\resizebox{0.9\textwidth}{!}{
\begin{tabular}{|l|l|l|}
\hline
\textbf{APPROACHES}  & \textbf{ORIGINAL CAPTIONS} & \textbf{ADVERSARIAL CAPTIONS}
\\ \hline 
% \hline
\UnTargetedAll & Two stuffed teddy bears sitting on a bed. & A blender that is sitting in the water.\\ 
                & A man riding a horse in front of a crowd. &   A bunch of food on a grill with meat being dogs. \\
                 & A person holding a hot dog on a bun. &  A close up view of an airplane with a knife. \\
                %   &  A fire hydrant with a hole in an oven. \\
                %   &  A group of baskets in a basket in a cage. \\

\hline
\TargetedFreq & A donut and a donut sitting on a table. & A plate with a doughnut and a donut on it. \\
                 & A man jumping a skateboard on a skateboard. &  A man jumping through the air with a skateboard. \\
                  & Two birds are flying over a building in a city. & Two birds sitting on a boat in the water. \\
                %   & A room with a couch and a chair on it. \\
                %     & A young child brushing his teeth in a car.\\
                    
\hline
\TargetedConf & A man riding a horse in front of a crowd. & A person riding a horse in front of a dog. \\
                 & A black and white photo of a city street with cars. &  A tower with a clock on top of it. \\
                  & A living room with a table and a table. & A man taking a picture in a bathroom mirror. \\
                %   &  A stuffed teddy bear is laying on the floor. \\
                %     & A counter and a plate of food on a table. \\
\hline
\UnTargetedFreq & Two cats sitting in a bath tub sink. & A black and white dog is standing in a boat. \\
                  & A black and white photo of a city street with cars. & A black and white photo of a city street with cars. \\
                    & A living room with a table and a table. & A living room with a couch and a table. \\
                    % &A train traveling down the tracks in the road. \\
                    % & A man taking a picture in a bathroom mirror. \\
\hline
\UnTargetedConf & A group of people walking around a parking meter. & A man is holding a parking meter on a pole.\\ 
                    & A television and a television in a room. & A living room with a couch and a chair. \\
                    & A vase with white flowers on a desk. & A vase with white flowers on a desk. \\
                    % & A blue plate of donuts on a table. \\
                    % & Two people standing in a living room playing a video game. \\
\hline
\end{tabular}}
\end{center}
\end{table*}

\section{Scales}

Table~\ref{tab:results_success_scale} shows the robustness of adversarial images generated using the proposed attacks against resizing with different scales. The targeted attacks are more robust in comparison with the non-targeted attacks since they add more perturbations to images to generate particular classes. These results show that the adversarial images are more robust for bigger scales in comparison with smaller scales.

\section{Histograms}
Figure~\ref{fig:histo_probs} shows the histograms of number of boxes and mean probabilities. The first row includes the histogram of the number of boxes, having the predicted label as the targeted class ($\classK$), with a positive IoU with the mask. It also includes the histogram of the mean probabilities of the targeted class for the boxes in the targeted attacks. The second row includes the histogram of the number of boxes with a positive IoU with the mask. It also includes the histogram of mean probabilities of the original class ($o_{pick}$) for the boxes in the non-targeted attacks. 
% This shows that the number of boxes for the attacks against the frequent object is greater than the attacks against the confident object.
\green{This confirms that as expected the total number of boxes for the frequent object is greater than the confident object.}
The mean probability of the targeted class for both \TargetedConf and \TargetedFreq are almost similar; however, the mean probability of the original class for \UnTargetedConf is more than \UnTargetedFreq.

Figure~\ref{fig:histo_iteration} shows the histogram of number of iterations. The first row shows the histogram of the number of iterations for the targeted attacks and the second row for the non-targeted attacks. The maximum number of iterations is $60$. If an attack takes $60$ iterations, this indicates an unsuccessful attack (we do not show the unsuccessful attacks for \TargetedConf when there is no bounding box having a positive IoU with the mask). The histograms show that attacking the most frequent object requires more iterations in comparison with attacking the most confident object. This is because attacking the most frequent object requires changing the label of more boxes in the image. As expected, the targeted attacks take more iterations than the non-targeted ones.

\section{Visualizing Adversarial Noise}
\green{We visualize the amount of noise added to several examples in Figure~\ref{fig:pert_targeted} and \ref{fig:pert_untargeted}. As shown, in each example, the noise is added to a region of image corresponding to the selected object ($\ro_{pick}$). We did this by defining $M$ (our binary mask) in Equation 2. For example, the sample objects of ``grass'' and ``giraffe'' are chosen and attacked in the first two rows of Figure~\ref{fig:pert_targeted} by \TargetedFreq. These objects are the most frequent objects in the corresponding images and their corresponding masks are bigger than the last two rows in the figure. They are also successfully changed to ``star'' and ``lion'', respectively. In the last two rows, the sample objects of ``kite'' and ``skateboard'' (the most confident objects) are attacked in the images by \TargetedConf. They are changed to ``eyes'' and ``column'', respectively. In addition, we have also shown some samples of our non-targeted attacks in Figure~\ref{fig:pert_untargeted}. For example, the sample objects of ``tie'' and ``bench'' are attacked in the first two rows of the figure by \UnTargetedFreq. These are the most frequent objects in the images. In the last two rows, ``whiskers'' and ``light'' are attacked by \UnTargetedConf. These are the most confident objects in the images. In these non-targeted attacks, we do not target any specific class and the most frequent objects usually come with bigger masks compared to the most confident ones.}

\section{Captions}
Table~\ref{tab:results_generated_captions} shows examples of predicted captions for adversarial and original images using different variants of \pickObjectAttack and \UnTargetedAll. As we can see \UnTargetedAll leads to image captions which are entirely unrelated with the original captions such as ``A man riding a horse in front of a crowd'' becomes ``A bunch of food on a grill with meat being dogs''.

\begin{figure*}[htbp]
\begin{center}
% \fbox{\rule{0pt}{2in} \rule{.9\linewidth}{0pt}}
\includegraphics[width=0.48\linewidth]{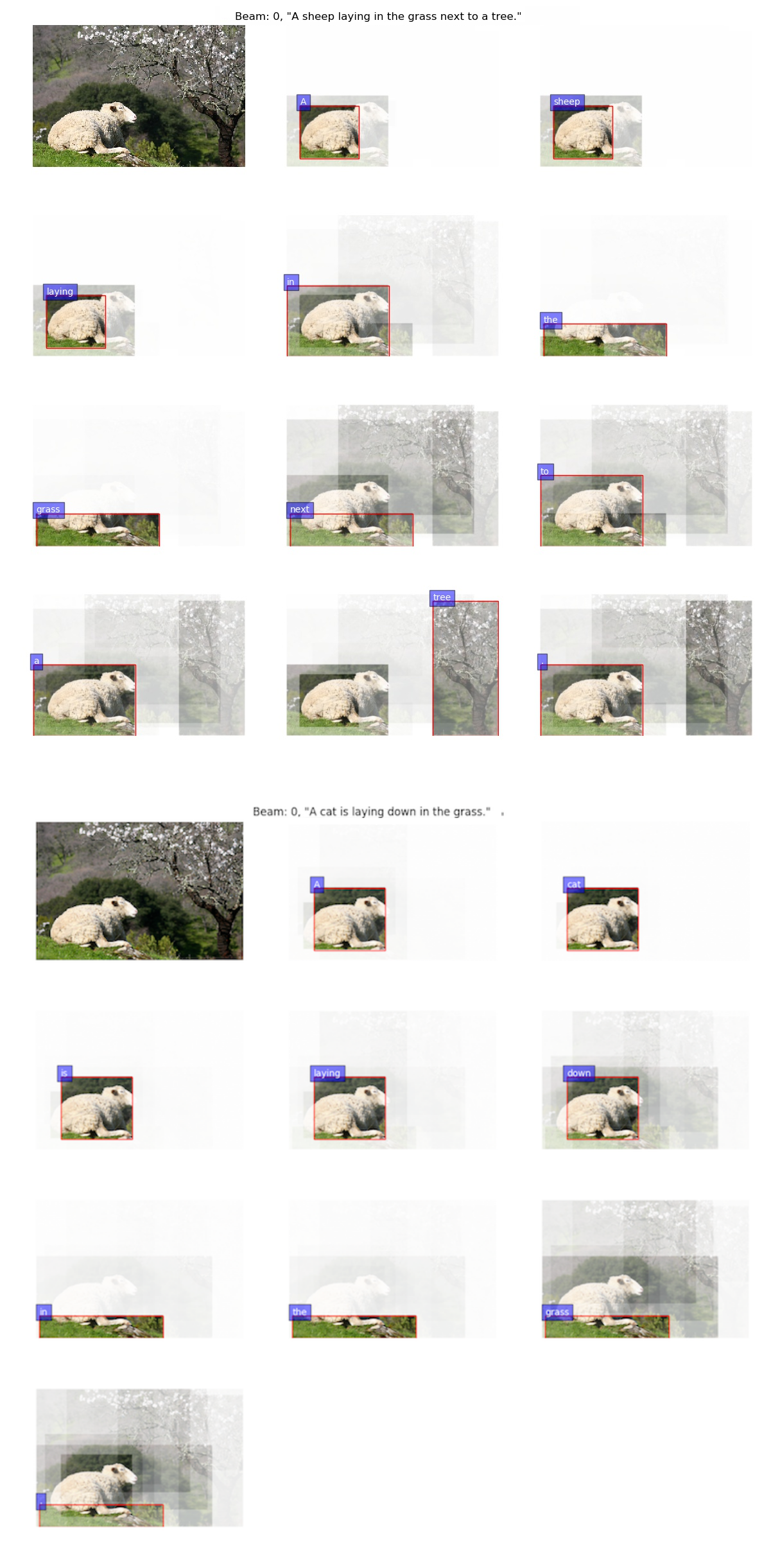}
\end{center}
  \caption{The attention visualization for the caption of the original image (the top image) and the adversarial image (the bottom image). The adversarial image was obtained by \TargetedConf and the original class was ``sheep'' ($o_{pick}$). In this example, the caption of the adversarial image includes the targeted class, ``cat'' ($k$).}
\label{fig:attention_correct}
\end{figure*}

\begin{figure*}[htbp]
\begin{center}
% \fbox{\rule{0pt}{2in} \rule{.9\linewidth}{0pt}}
\includegraphics[width=0.52\linewidth]{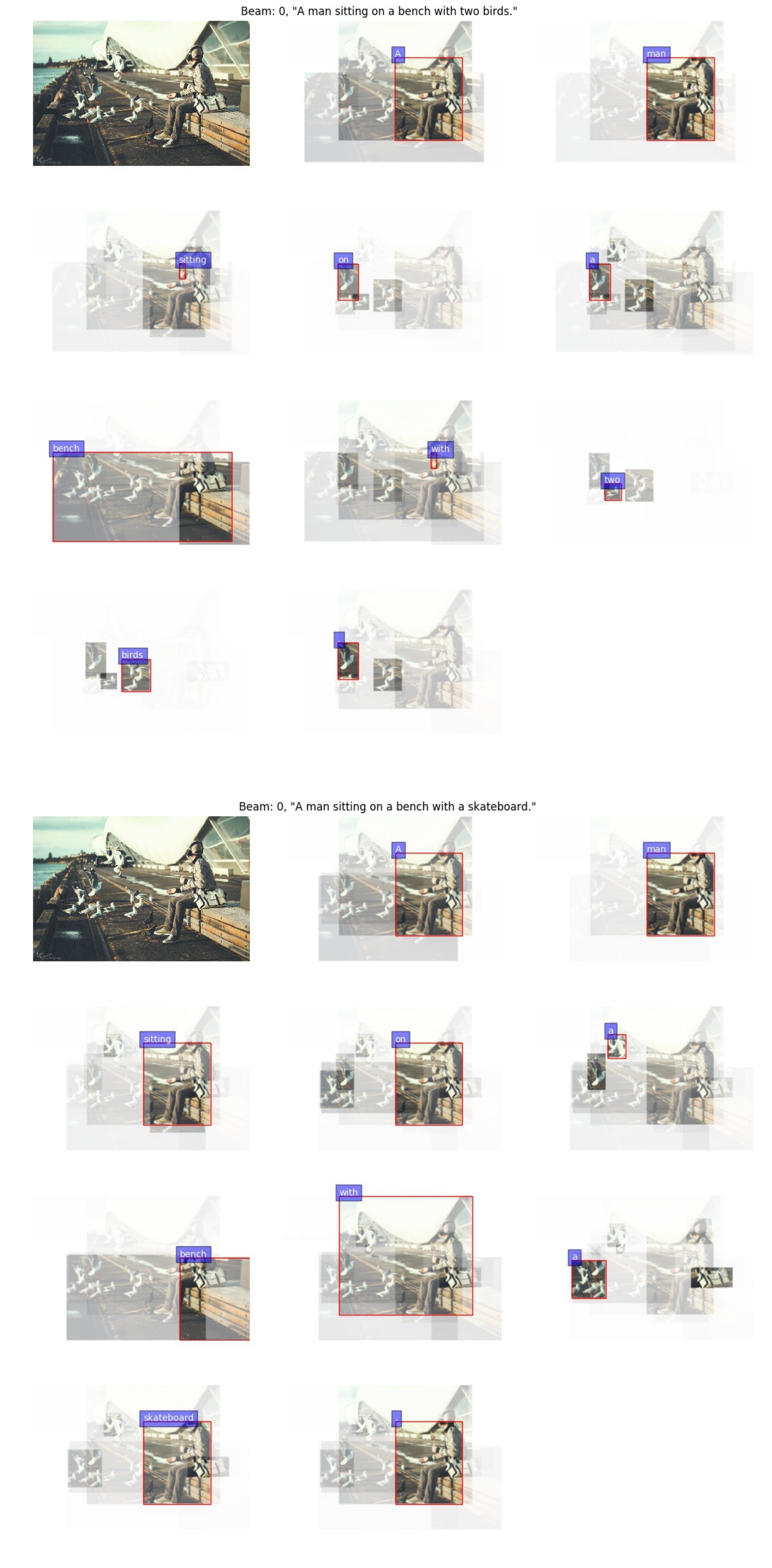}
\end{center}
  \caption{The attention visualization for the caption of the original image (the top image) and the adversarial image (the bottom image). The adversarial image was obtained by \TargetedFreq and the original class was ``bird'' ($o_{pick}$). In this example, the caption of the adversarial image does not include the targeted class, ``sign'' ($k$).}
\label{fig:attention_wrong}
\end{figure*}

\end{document}